\documentclass[10pt,twocolumn,letterpaper]{article}

\usepackage{cvpr}              % To produce the CAMERA-READY version
\usepackage{graphicx}
\usepackage{amsmath}
\usepackage{amssymb}
\usepackage{booktabs}

\usepackage{multirow}
\usepackage{lipsum}
\usepackage{bbding}

\usepackage[pagebackref,breaklinks,colorlinks]{hyperref}

\usepackage[capitalize]{cleveref}
\crefname{section}{Sec.}{Secs.}
\Crefname{section}{Section}{Sections}
\Crefname{table}{Table}{Tables}
\crefname{table}{Tab.}{Tabs.}

\def\confName{CVPR}
\def\confYear{2022}

\newcommand{\myparagraph}[1]{{\vspace{.5em} \noindent \bf #1}}

\begin{document}

\title{Point RCNN: An Angle-Free Framework for Rotated Object Detection}

\author{Qiang Zhou\footnotemark[1], ~~~~~~Chaohui Yu\footnotemark[1], ~~~~~~Zhibin Wang\footnotemark[2], ~~~~~~Hao Li\\
Alibaba Group\\
{\tt\small \{jianchong.zq,huakun.ych,zhibin.waz,lihao.lh\}@alibaba-inc.com}
}

%%%%% Sexy front cover
\makeatletter
\let\@oldmaketitle\@maketitle% Store \@maketitle
\renewcommand{\@maketitle}{\@oldmaketitle% Update \@maketitle to insert...
 \centering
    \includegraphics[width=0.85\linewidth]{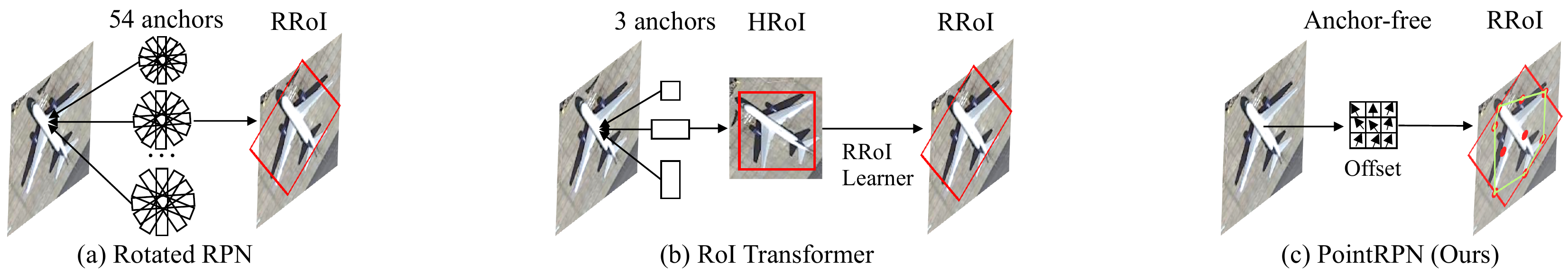}
    \vspace{-.1in}
    \captionof{figure}{
    Comparison of different methods for generating rotated RoI (RRoI). 
    (a) Rotated RPN places multiple rotated anchors with different angles, scales, and aspect ratios.
    (b) RoI Transformer proposes a RRoI learner to model the RRoI from the horizontal RoI (HRoI) for each feature point based on 3 anchors.
    (c) Our proposed PointRPN generates accurate RRoI in an anchor-free and angle-free manner.
    }
    \label{fig:frontcover}
    %\vspace{-.05in}
  \bigskip}
\makeatother

\maketitle

\renewcommand{\thefootnote}{\fnsymbol{footnote}}
\footnotetext[1]{These authors contributed equally to this work.}
\footnotetext[2]{Corresponding author.}

%%%%%%%%% ABSTRACT
\begin{abstract}

Rotated object detection in aerial images is still challenging due to arbitrary orientations, large scale and aspect ratio variations, and extreme density of objects. 
Existing state-of-the-art rotated object detection methods mainly rely on angle-based detectors. 
However, angle regression can easily suffer from the long-standing boundary problem. 
To tackle this problem, we propose a purely angle-free framework for rotated object detection, called Point RCNN, which mainly consists of PointRPN and PointReg. 
In particular, PointRPN generates accurate rotated RoIs (RRoIs) by converting the learned representative points with a coarse-to-fine manner, which is motivated by RepPoints. 
Based on the learned RRoIs, PointReg performs corner points refinement for more accurate detection. 
In addition, aerial images are often severely unbalanced in categories, and existing methods almost ignore this issue. 
In this paper, we also experimentally verify that re-sampling the images of the rare categories will stabilize training and further improve the detection performance. 
Experiments demonstrate that our Point RCNN achieves the new state-of-the-art detection performance on commonly used aerial datasets, including DOTA-v1.0, DOTA-v1.5, and HRSC2016.

\end{abstract}

%%%%%%%%% BODY TEXT

\section{Introduction}

%Deep neural networks~\cite{krizhevsky2012imagenet,he2016deep} have significantly improved the performance of diverse computer vision applications, \eg, image classification and object detection. 
%
Although object detection has achieved significant progress in natural images, it still remains challenging for rotated object detection in aerial images, due to the arbitrary orientations, large scale and aspect ratio variations, and extreme density of objects~\cite{xia2018dota}.
Rotated object detection aims at predicting a set of oriented bounding box (OBB) and the corresponding classes in an aerial image, which has been serving as an essential step in many applications, \eg, urban management, emergency rescue, precise agriculture~\cite{ding2021object}. 
Modern rotated object detectors can be divided into two categories in terms of the representation of OBB: angle-based detectors and angle-free detectors.

In angle-based detectors, an OBB of a rotated object is usually represented as a five-parameter ($x, y, w, h, \theta$). Most existing state-of-the-art methods are angle-based detectors relying on two-stage RCNN frameworks~\cite{jiang2018r,ding2019learning,yang2019scrdet,han2021redet,xie2021oriented}. Generally, these methods use an RPN to generate horizontal or rotated RoIs, then use a designed RoI pooling operator to extract features from these RoIs. Finally, an RCNN head is used to predict the OBB and the corresponding classes. 
Compared to two-stage detectors, one-stage angle-based detectors~\cite{ma2018arbitrary,zhang2018toward,yang2020arbitrary,yang2019r3det,han2021align} directly regress the OBB and classify them based on dense anchors for efficiency.

However, angle-based detectors usually introduce a long-standing boundary discontinuity problem~\cite{DCL,yang2020scrdet++} due to the periodicity of angle and exchange of edges. Moreover, the unit between ($x, y, w, h$) and angle $\theta$ of the five-parameter representation is not consistent. These obstacles will cause the training unstable and limit the performance. 

In contrast with angle-based detectors, angle-free detectors usually represent a rotated object as an eight-parameter OBB ($x_1, y_1, x_2, y_2, x_3, y_3, x_4, y_4$), which denotes the four corner points of a rotated object.
Modern angle-free detectors~\cite{azimi2018towards,qian2019learning,xu2020gliding,yi2021oriented} directly perform quadrilateral regression, which is more straightforward than the angle-based representation. Unfortunately, although abandoning angle regression and the parameter unit is consistent as well, the performance of existing angle-free detectors is still relatively limited.
How to design a more straightforward and effective framework to alleviate the boundary discontinuity problem is the key to the success of rotated object detectors.

In this paper, we propose a purely angle-free framework for rotated object detection, called Point RCNN, which can alleviate the boundary discontinuity problem and attain state-of-the-art performance. Concretely, Point RCNN is a two-stage detector and mainly consists of an RPN (PointRPN) and an RCNN head (PointReg), which are both angle-free.
%PointRPN serves as a RPN network. 
Given an input feature map, first, PointRPN learns a set of representative points for each feature point with a coarse-to-fine manner.
Then, rotated RoI (RRoI) is generated through the $\tt{MinAreaRect}$ function of OpenCV~\cite{opencv_library}.
%
%PointReg serves as an angle-free RCNN head. Given a RRoI, 
Finally, PointReg applies a rotate RoI Align~\cite{han2021redet, ding2019learning} operator to extract RRoI features, and then refines and classifies the eight-parameter OBB of corner points.
%
%\textcolor{yellow}{In addition, we propose a balanced dataset strategy to rebalance the long-tailed dataset to stabilize convergence during training.}
In addition, the existing methods almost ignore the category imbalance in aerial images, and we propose to resample images of rare categories to stabilize convergence during training.

The contributions of this paper are as follows:
\begin{itemize}
\item We propose Point RCNN, a purely angle-free framework for rotated object detection. Without introducing angle prediction, Point RCNN is able to address the boundary discontinuity problem.

\item We propose PointRPN and PointReg to reformulate angle prediction as the more straightforward points regression. Both of them are angle-free and have consistent parameter units. 
%\textcolor{yellow}{We further propose a balanced dataset strategy to stabilize the convergence process.}
We further propose to resample images of rare categories to stabilize training and improve overall performance.

\item Compared with the state-of-the-art methods, our Point RCNN framework attains higher detection performance on several large-scale datasets.
\end{itemize}

\section{Related Work}

\subsection{Horizontal Object Detection}

In the past decade, object detection has become an important computer vision task and has received considerable attention.
One line of research focuses on two-stage detectors~\cite{girshick2014rich,girshick2015fast,ren2015faster,he2017mask,lin2017feature,cai2018cascade,hu2018relation}, which first generates a sparse set of Regions of Interests (RoI) with a Region Proposal Network (RPN), and then performs classification and bounding box regression. 
While two-stage detectors still attract much attention, another line of research tends to develop efficient one-stage detectors~\cite{liu2016ssd,redmon2016you,lin2017focal,law2018cornernet,tian2019fcos,duan2019centernet,yang2019reppoints}, in which SSD~\cite{liu2016ssd} and YOLO~\cite{redmon2016you} are the fundamental methods that use a set of pre-defined anchor boxes to predict object category and anchor box offsets.
Recently, some anchor-free methods~\cite{law2018cornernet,duan2019centernet,yang2019reppoints} detect object by predicting the center or corner or representative points, which also inspire us to develop the angle-free detector for rotated object.

\subsection{Rotated Object Detection}

In terms of the representation of oriented bounding box (OBB), modern rotated object detectors can be mainly divided into two categories: angle-based detectors and angle-free detectors. 

%\noindent \textbf{Angle-based detectors} 
\myparagraph{Angle-based detectors} detect rotated object by learning a five-parameter OBB ($x, y, w, h, \theta$), in which ($x, y, w, h$) denotes a horizontal bounding box and $\theta$ denotes the rotated angle between the longer edge and the horizontal axis.
RRPN~\cite{ma2018arbitrary} and R$^{2}$PN~\cite{zhang2018toward} make use of multiple rotated anchors with different angles, scales, and aspect ratios, which improves the performance while increasing the computational complexity (see~\cref{fig:frontcover}(a)). 
R$^2$CNN~\cite{jiang2018r} proposes to detect horizontal and rotated bounding box simultaneously with multi task learning.
RoI Transformer~\cite{ding2019learning} proposes a rotated RoI (RRoI) learner to transform a horizontal RoI into a RRoI, which provides more accurate RRoIs with a complex pipeline (see~\cref{fig:frontcover}(b)).
SCRDet~\cite{yang2019scrdet} enhances features with attention module and proposes an IoU-smooth $L_1$ loss to alleviate the loss discontinuity issue.
%SCRDet++~\cite{yang2020scrdet++} extends SCRDet with image-level and instance-level denoising modules to enhance the detection to small and cluttered objects.
CSL~\cite{yang2020arbitrary} reformulates angle prediction from regression to classification to alleviate discontinuous boundary problem.
GWD~\cite{GWD2021} and KLD~\cite{KLD2021} propose more efficient loss function for OBB regression.
%R3Det~\cite{yang2019r3det} proposes a refined single-stage rotation detector for fast and accurate object detection by using a progressive regression approach from coarse to fine granularity.
S$^{2}$A-Net~\cite{han2021align} proposes a single-shot alignment network to realize full feature alignment and alleviates the inconsistency between regression and classification.
Recently, ReDet~\cite{han2021redet} proposes to use rotation-equivariant network to encode rotation equivariance explicitly and presents rotation-invariant RoI Align to extract rotation-invariant features.
Oriented R-CNN~\cite{xie2021oriented} proposes a two-stage detector that consists of an oriented RPN for generating RRoI and an oriented RCNN for refining the RRoI. 
Both ReDet and Oriented RCNN provide promising accuracy.

However, the boundary problem in the angle regression learning still causes training unstable and limits the performance.
While angle-based detectors still find many applications, angle-free methods are getting more and more attention from the community.

\begin{figure*}[t!]
\centering 
\includegraphics[width=0.92\linewidth]{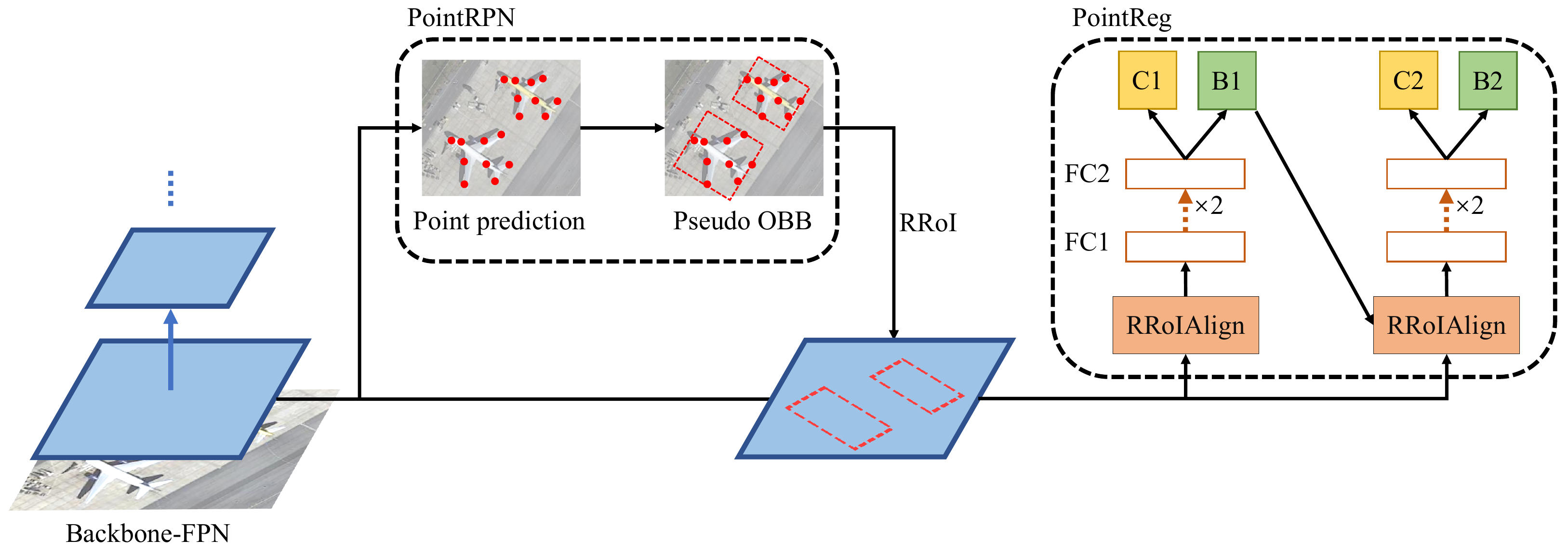}
\vspace{-.1in}
\caption{The proposed angle-free Point RCNN framework for rotated object detection. Point RCNN mainly consists of two modules: PointRPN for generating rotated proposals, and PointReg for refining for more accurate detection. ``RRoI'' denotes rotated RoI, ``FC'' denotes fully-connected layer, ``C'' and ``B'' represent the predicted category and rotated box coordinates of each RRoI, respectively.}
\vspace{-.1in}
\label{fig:pointrcnn}
\end{figure*}

%\noindent \textbf{Angle-free detectors} 
\myparagraph{Angle-free detectors} reformulate rotated object regression as learning a eight-parameter OBB ($x_1, y_1, x_2, y_2, x_3, y_3, x_4, y_4$), which represents the four corner points of a rotated object.
%Textboxes++~\cite{liao2018textboxes++} directly predicts arbitrary-oriented word bounding boxes via a regression model by quadrilateral representation.
ICN~\cite{azimi2018towards} proposes to directly estimate the four vertices of a quadrilateral to regress an oriented object based on image pyramid and feature pyramid.
%RSDet~\cite{qian2019learning} proposes a modulated rotation loss to further solve the problem of inconsistent parameter regression.
RSDet~\cite{qian2019learning} and Gliding Vertex~\cite{xu2020gliding} achieve more accurate rotated object detection via directly quadrilateral regression prediction.
Recently, BBAVectors~\cite{yi2021oriented} extends the horizontal keypoint-based object detector to the oriented object detection task.
CFA~\cite{BeyondBBox} proposes a convex-hull feature adaptation approach for configuring convolutional features.
Compared to angle-based methods, angle-free detectors are more straightforward and can alleviate the boundary problem to a large extent. However, the performance is relatively limited yet.

In this paper, we propose an effective angle-free framework for rotated object detection, \ie, Point RCNN, which mainly consists of PointRPN and PointReg. Compared with other RRoI generation methods, our PointRPN generates accurate RRoI in an anchor-free and angle-free manner (see~\cref{fig:frontcover}(c)).

\section{Point RCNN}

%We introduce a purely angle-free framework for rotated object detection---Point RCNN, which consists of a Point RPN and a Point RCNN head. Point RCNN is able to alleviate the long-standing boundary problem by reformulating angle prediction as the more straightforward points regression.

The overall structure of our Point RCNN is depicted in Fig.~\ref{fig:pointrcnn}.
We start by revisiting the boundary discontinuity problem of angle-based detectors.
Then, we describe the overall pipeline of Point RCNN. % which consists of PointRPN and a PointReg. 
Finally, we elaborate the PointRPN and PointReg modules, and propose a balanced dataset strategy to rebalance the long-tailed datasets during training. %\eg, DOTA~\cite{xia2018dota}.

\subsection{Boundary Discontinuity}
\label{med:boundary_issue}

% \begin{figure}[t!]
% \centering 
% \includegraphics[width=0.6\linewidth]{figures/boundary2.pdf}
% \caption{Loss discontinuity of corner points regression. Red and yellow indicate two similar corner point sets, a slight rotation ($\pm1^{\circ}$) will change the order of the corner points.}
% \label{fig:boundary2}
% \end{figure}

% \begin{figure}[t!]
%      \centering
%      \begin{subfigure}[b]{0.45\linewidth}
%          \centering
%          \includegraphics[width=\linewidth]{figures/boundary.pdf}
%          \caption{Boundary discontinuity of angle regression}
%          \label{fig:boundary}
%      \end{subfigure}
%      %\hfill
%      \quad
%      \begin{subfigure}[b]{0.45\linewidth} 
%          \centering
%          \includegraphics[width=\linewidth]{figures/boundary2.pdf}
%          \caption{Loss discontinuity of corner points regression}
%          \label{fig:boundary2}
%      \end{subfigure}
%     \caption{Discontinuity problem in angle prediction and corner points regression. a) Red and yellow indicate two different targets. Although the two square-like targets have slightly different edge lengths, there is a huge gap between the angles $\theta$. b) Red and yellow indicate two similar corner point sets, a slight rotation ($\pm1^{\circ}$) will change the order of the corner points.}
%     \label{fig:boundary_discontinuity}
% \end{figure}

\begin{figure}[t!]
\centering 
\includegraphics[width=0.5\linewidth]{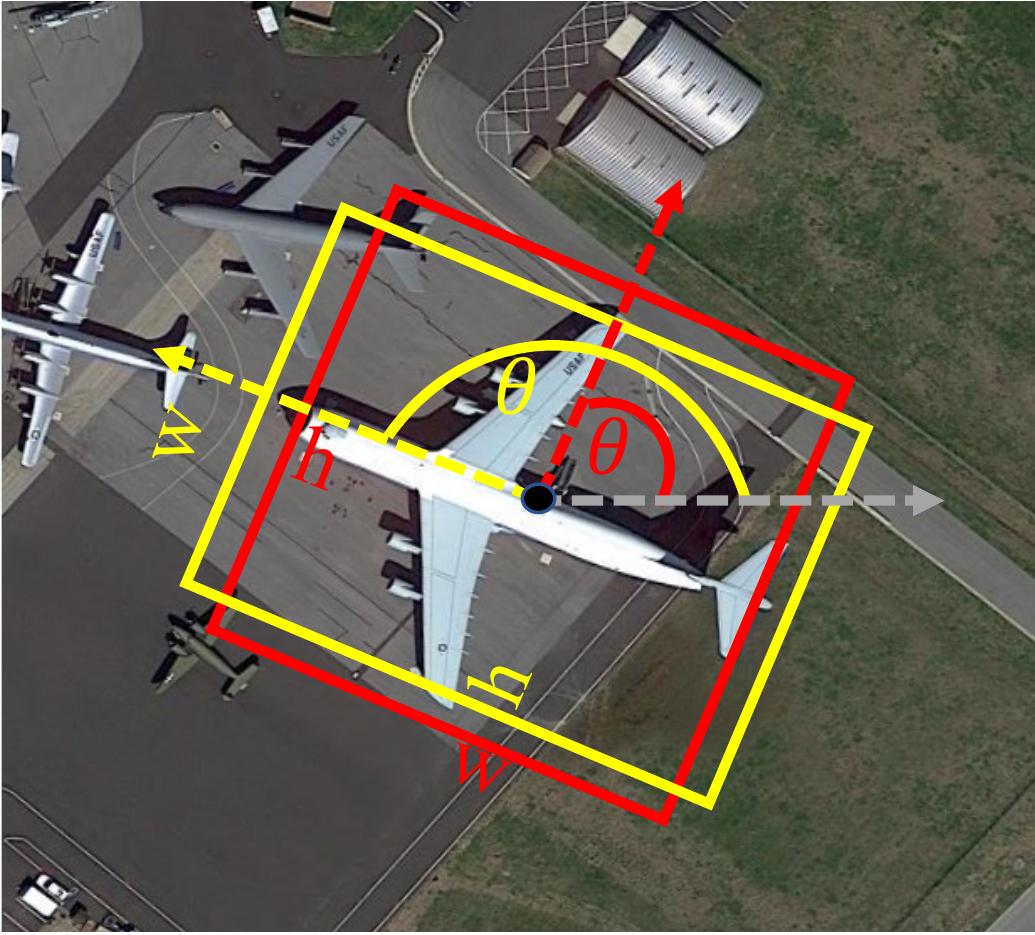}
\vspace{-.1in}
\caption{Boundary discontinuity problem of angle prediction. Red and yellow indicate two different targets.
%Although the two square-like targets have slightly different edge lengths, there is a huge gap between the angles $\theta$.
}
\label{fig:boundary}
\vspace{-.2in}
\end{figure}

Boundary problem~\cite{DCL,yang2020scrdet++} is a long-standing problem that existed in angle-based detectors. 
Take the commonly used five-parameter OBB representation $(x, y, w, h, \theta)$ as an example, where $(x, y)$ represents the center coordinates, $(w, h)$ represents the shorter and longer edges of the box, and $\theta$ represents the angle between the longer edge and the horizontal axis.
As shown in Fig.~\ref{fig:boundary}, when the target box is approximately square, a slight variation in edge length may cause $w$ and $h$ to swap, leading to a substantial variation of $\pi/2$ in angle $\theta$.

This boundary discontinuity issue in angle prediction will confuse the optimization of the network and limit the detection performance.

\subsection{Overview}
\label{med:overview}

%To tackle the boundary problem in angle regression, in this paper, we propose a straightforward and efficient angle-free framework for rotated object detection. Instead of predicting the angle like many previous angle-based two-stage methods~\cite{ding2019learning,han2021redet,xie2021oriented}, our proposed Point RCNN reformulates the oriented bounding box (OBB) task as learning the representative points of object in the RPN phase and modeling the corner points in the RCNN refine phase, which is totally angle-free.

The overall pipeline of Point RCNN is shown in Fig.~\ref{fig:pointrcnn}.
During training, Backbone-FPN first extracts feature maps given an input image.
%\textcolor{yellow}{with five levels ($P_2$, $P_3$, $P_4$, $P_5$, $P_6$) (to remove)}. 
Then, PointRPN performs representative points regression and generates pseudo OBB for rotated RoI (RRoI). 
Finally, for each RRoI, PointReg refines the corner points and classifies them for final detection results.
%\textcolor{yellow}{Besides, we propose a balanced dataset strategy to overcome the imbalanced dataset issue during training.} 
Besides, we propose to resample images of rare categories to stabilize training and improve the overall performance.

The overall training objective is described as:
\begin{equation}
    \mathcal{L} = \mathcal{L}_{PointRPN} + \mathcal{L}_{PointReg},
\end{equation}
where $\mathcal{L}_{PointRPN}$ denotes the losses in PointRPN, and $\mathcal{L}_{PointReg}$ denotes the losses in PointReg.
We will describe them in detail in the following sections.

\subsection{PointRPN}
\label{med:pointrpn}

\begin{figure}[t!]
\centering 
\includegraphics[width=0.9\linewidth]{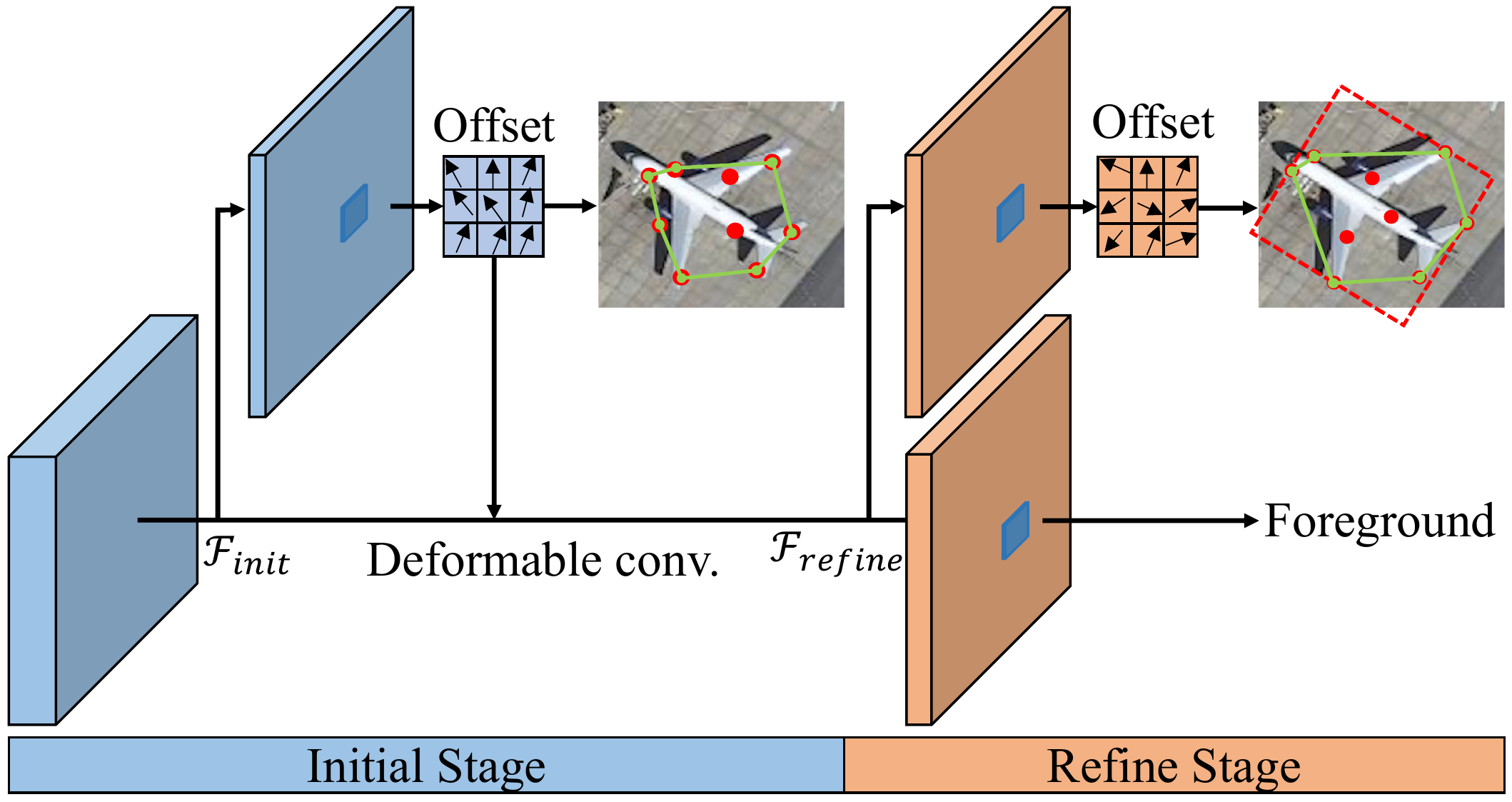}
\vspace{-.1in}
\caption{The structure of the proposed PointRPN. The red points are the learned representative points, and the green polygon represents the converted convex-hull. The red dotted OBB is converted from the representative points with the $\tt{MinAreaRect}$ function of OpenCV~\cite{opencv_library} for generating RRoI.}
\label{fig:pointrpn}
\vspace{-.1in}
\end{figure}

Existing rotated object detection methods generate rotated proposals indirectly by transforming the outputs of RPN~\cite{faster-rcnn} and suffer from the boundary discontinuity problem caused by angle prediction.
For example,~\cite{han2021redet,ding2019learning} use RoI transformer to convert horizontal proposals to rotated proposals with an additional angle prediction task.
Unlike these methods, in this paper, we propose to directly predict the rotated proposals with representative points learning. 
The learning of points is more flexible, and the distribution of points can reflect the angle and size of the rotated object. The boundary discontinuity problem can thus be alleviated without angle regression.

%Unlike these methods, in this paper, we propose to directly predict the rotated proposals without angle regression. 

\myparagraph{Representative Points Prediction.}
Inspired by RepPoints~\cite{yang2019reppoints} and CFA~\cite{BeyondBBox}, we propose PointRPN to predict the representative points in the RPN stage.
The predicted points can effectively represent the rotating box and can be easily converted to rotated proposals in subsequent RCNN stages.

As shown in Fig.~\ref{fig:pointrpn}, PointRPN learns a set of representative points for each feature point.
In order to make the features better adapt to the representative points learning, we take a coarse-to-fine prediction manner. In this way, 
the features will be refined with DCN~\cite{dai2017deformable} and the predicted offset in the initial stage. %as depicted in \cref{fig:pointrpn}.
For each feature point, the predicted representative points of the two stages are:
\begin{equation}
    \begin{split}
        \mathcal{R}^{init} &= \{(x^0_i+\Delta x_{i}^0, y^0_i+\Delta y_i^0)\}^{K}_{i=1}, \\
        \mathcal{R}^{refine} &= \{(x_i^1 + \Delta x_i^1, y_i^1 + \Delta y_i^1)\}^{K}_{i=1},
    \end{split}
\end{equation}
where $K$ denotes the number of predicted representative points and we set $K=9$ by default. 
$\{(x^0_i, y^0_i)\}^{K}_{i=1}$ denotes the initial location, 
$\{(\Delta x_i^0, \Delta y_i^0)\}^{K}_{i=1}$ denote the learned offsets in the initial stage, and $\{(\Delta x_i^1, \Delta y_i^1)\}^{K}_{i=1}$ denote the learned offsets in the refine stage.

%In horizontal object detection, the horizontal bounding box (HBB) of a set of representative points $\mathcal{R}$ can be obtained by a conversion function, \eg, min-max or moment-based function in RepPoints~\cite{yang2019reppoints}. 
%However, in rotated object detection, it is challenging to adapt the representative points appropriately when using HBB as the learning target. %due to the arbitrary orientations, scale variations, extremely nonuniform object densities and large aspect ratios. 
% 
%To make the representative points adapt the rotated object more accurately, we take advantage of the OBB annotations.
%Thus a conversion method is needed to transform the learned points, so as to be supervised by the OBB annotations. 
%
%We follow CFA~\cite{BeyondBBox} to calculate the convex-hull $\mathcal{C}$ of the representative points with Jarvis March algorithm~\cite{jarvis1973identification}, denoted as $\Gamma(\cdot)$. The convex-hull $\mathcal{C}$ can be formulated as: $\mathcal{C} = \Gamma(\mathcal{R})$.

\myparagraph{Label Assignment.}
PointRPN predicts representative points for each feature point in the initial and refine stages. This section will describe how we determine the positive samples among all feature points for these two stages.

For the initial stage, we project each ground-truth box to the corresponding feature level according to its area, and then select the feature point closest to its center as the positive sample.
The rule used for projecting the ground-truth box $b_i^*$ to the corresponding feature level is defined as:
\begin{equation}
    l_i = \log_{2}{\left ( \sqrt{\frac{w_i h_i}{s} }  \right ) },
\end{equation}
where $s$ is a hyper-parameter and is set to 16 by default. $w_i$ and $h_i$ are the width and height of $b_i^*$.
%\textcolor{blue}{The calculated $l_i$ will be further limited to the range of $[3, 7]$, since we make predictions for the five feature levels of ($P_3$, $P_4$, $P_5$, $P_6$, $P_7$).}

For the refine stage, we use the predicted representative points from the initial stage to help determine the positive samples.
Specifically, for each feature point with its corresponding prediction $\mathcal{R}^{init}$, if the maximum convex-hull IoU (defined in~\cref{eq:ciou}) between $\mathcal{R}^{init}$ and ground-truth boxes exceeds the threshold $\tau$, we select this feature point as a positive sample. We set $\tau$ = 0.1 in all our experiments.

\myparagraph{Optimization.}
The optimization of the proposed PointRPN is driven by classification loss and rotated object localization loss. The learning objective is formulated as:
\begin{equation}
    \mathcal{L}_{PointRPN} = \lambda_1 ~^{+}\mathcal{L}^{init}_{loc} + \lambda_2 ~\mathcal{L}^{refine}_{cls} + \lambda_3 ~^{+}\mathcal{L}^{refine}_{loc},
\label{eq:pointrpnloss} 
\end{equation}
where $\lambda_1, \lambda_2$, and $\lambda_3$ are the trade-off parameters and are set to 0.5, 1.0, and 1.0 by default, respectively.
$^{+}\mathcal{L}^{init}_{loc}$ denotes the localization loss of the initial stage. 
$\mathcal{L}^{refine}_{cls}$ and $^{+}\mathcal{L}^{refine}_{loc}$ denote the classification loss and localization loss of the refine stage. 
Note that the classification loss is only calculated in the refine stage, and the two localization losses are only calculated for the positive samples.

In the initial stage, the localization loss is conducted between the convex-hulls converted from the learned points $\mathcal{R}^{init}$ (see initial stage in Fig.~\ref{fig:pointrpn}) and the ground-truth OBBs. 
We use convex-hull GIoU loss~\cite{BeyondBBox} to calculate the localization loss:
\begin{equation}
    ^{+}\mathcal{L}^{init}_{loc} = \frac{1}{N^{0}_{pos}} \sum_{i} \Big(1 - \mathrm{CIoU}\big(\Gamma(\mathcal{R}^{init}_i), \Gamma(b_{i}^*) \big)  \Big),
\label{eq:init_loc} 
\end{equation}
where $N^{0}_{pos}$ indicates the number of positive samples of the initial stage. $b_{i}^{*}$ is the matched ground-truth OBB. $\mathrm{CIoU}$ represents the convex-hull GIoU between the two convex-hulls $\Gamma(\mathcal{R}^{init}_i)$ and $\Gamma(b_i^*)$, which is differential and can be calculated as:
\begin{equation}
    \begin{split}
    \mathrm{CIoU} \big( \Gamma(\mathcal{R}^{init}_i), & \Gamma(b_i^*) \big) = \frac{\big| \Gamma(\mathcal{R}^{init}_i) \cap \Gamma(b_i^*) \big|}{\big| \Gamma(\mathcal{R}^{init}_i) \cup \Gamma(b_i^*) \big|} \\
    &- \frac{\big| \mathcal{P}_i \setminus \big(\Gamma(\mathcal{R}^{init}_i) \cup \Gamma(b_i^*) \big) \big| }{\mathcal{P}_i},
    \end{split}
\label{eq:ciou} 
\end{equation}
where the first term denotes the convex-hull IoU, and $\mathcal{P}_i$ denotes the smallest enclosing convex object area of $\Gamma(\mathcal{R}^{init}_i)$ and $\Gamma(b_i^*)$.
$\Gamma(\cdot)$ denotes the Jarvis March algorithm~\cite{jarvis1973identification} used to calculate the convex-hull from points.

The learning of the refine stage, which is responsible for outputting more accurate rotated proposals, is driven by both classification loss and localization loss. $\mathcal{L}^{refine}_{cls}$ is a standard focal loss~\cite{lin2017focal}, which can be calculated as:
\begin{equation}
    \mathcal{L}^{refine}_{cls} = \frac{1}{N^{1}_{pos}} \sum_{i} \mathrm{FL}(p_i, c_{i}^*),
\label{eq:FL} 
\end{equation}
\begin{equation}
\mathrm{FL}(p_i, c_{i}^*) = 
\begin{cases}
-\alpha (1 - p_i)^\gamma ~ \text{log} (p_i), & \text{if}~ c_i^* > 0;  \\ 
-(1 - \alpha) p_i^\gamma~ \text{log} (1 - p_i), & \text{otherwise},
\end{cases} 
\end{equation}
where $N^{1}_{pos}$ denotes the number of positive samples in the refine stage, $p_i$ and $c_i^*$ are the classification output and the assigned ground-truth category, respectively. $\alpha$ and $\gamma$ are hyper-parameters and are set to 0.25 and 2.0 by default. The localization loss $\mathcal{L}^{refine}_{loc}$ is similar to~\cref{eq:init_loc} and can be formulated as:
\begin{equation}
    ^{+}\mathcal{L}^{refine}_{loc} = \frac{1}{N_{pos}^1} \sum_{i} \Big(1 - \mathrm{CIoU}\big(\Gamma(\mathcal{R}^{refine}_i), \Gamma(b_{i}^*) \big)  \Big).
\label{eq:refine_loc} 
\end{equation}
With the refined representative points, pseudo OBB is converted using the $\tt{MinAreaRect}$ function of OpenCV~\cite{opencv_library}, which is then used for generating RRoI for PointReg.

As illustrated in Fig.~\ref{fig:vis_rpn}, our PointRPN can automatically learn extreme points and semantic key points of rotated objects.

\subsection{PointReg}
\label{med:pointreg}

\begin{figure}[t!]
\centering 
\includegraphics[width=0.95\linewidth]{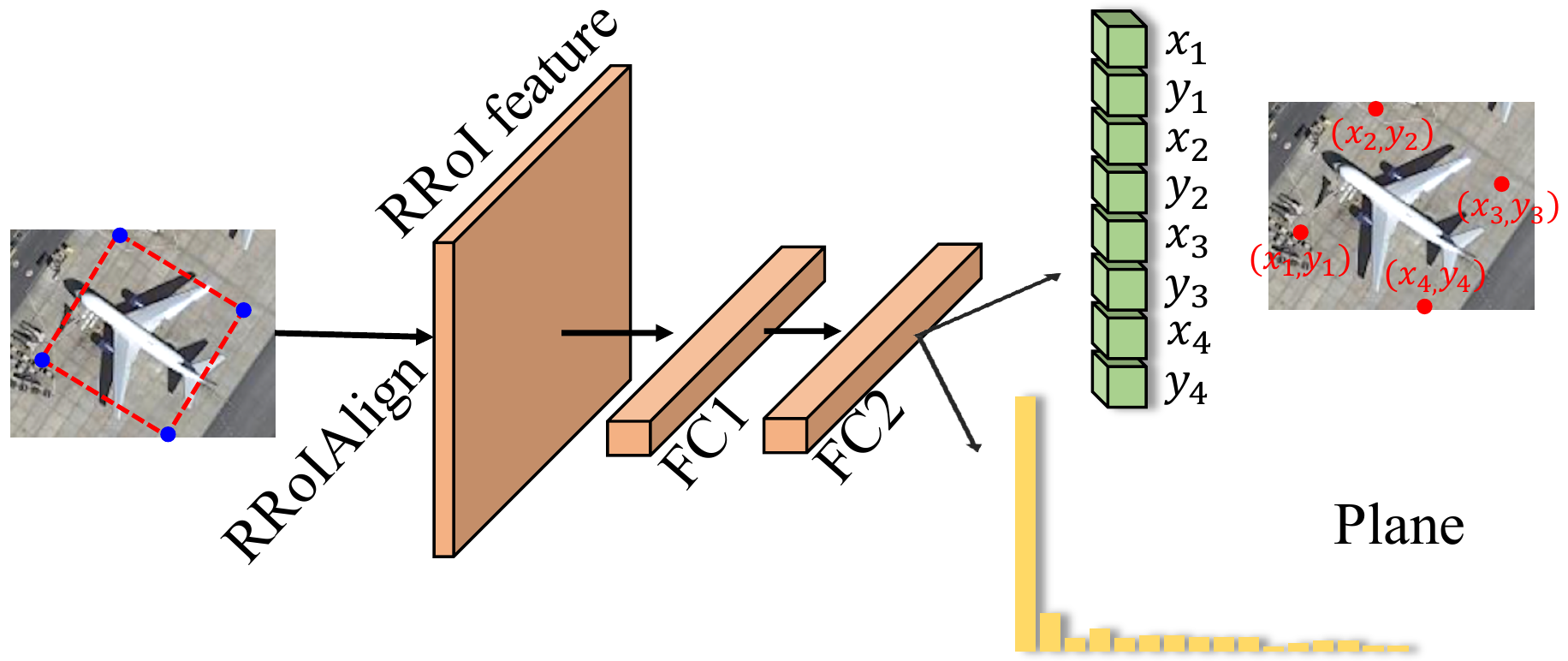}
\vspace{-.1in}
\caption{The diagram of the proposed PointReg. For simplicity, we only show the first stage of PointReg. Blue and red points denote the four corner points of the input RRoI and the refined results, respectively.
}
\label{fig:pointreg}
\vspace{-.1in}
\end{figure}

\myparagraph{Corner Points Refine.}
The rotated proposals generated by PointRPN already provide a reasonable estimate for the target rotated objects.
To avoid the problems caused by angle regression and further improve the performance, we turn to refine the four corners of the rotated proposals in the RCNN stage.
As shown in \cref{fig:pointreg}, with the rotated proposals as input, we use a RRoI feature extractor~\cite{ding2019learning,han2021redet} to extract RRoI features.
Then, given the RRoI features, two consecutive fully-connected and ReLU layers are used to encode the RRoI features. Finally, two fully-connected layers are responsible for predicting the class probability $P$ and refined corners $\mathcal{C}$ of the corresponding rotated object. 
The refined corners are represented as:
\begin{equation}
    \mathcal{C} = \{(x_i+\Delta x_{i}, y_i+\Delta y_i)\}^{4}_{i=1},
    \label{eq:corner_offset}
\end{equation}
where $\{(x_i, y_i)\}^{4}_{i=1}$ denotes the corner coordinates of the input rotated proposals. $\{(\Delta x_{i}, \Delta y_{i})\}^{4}_{i=1}$ denotes the predicted corner offsets.

Instead of directly performing angle prediction, we refine the four corners of the input rotated proposals. There are three advantages of adopting corner points refinement: 
1).~it can alleviate the boundary discontinuity problem caused by angle prediction;
2).~the parameter units are consistent among the eight parameters;
3).~it is possible to improve the localization accuracy using a coarse-to-fine manner.

We can easily extend PointReg to cascade structure for better performance. As shown in \cref{fig:pointrcnn}, in the cascade structure, the refined rotated proposals of the previous stage are used as the input of the current stage.

%Given the rotated RoI (RRoI) of the proposed angle-free PointRPN, we introduce our RCNN head, \ie, PointReg, which is also angle-free.
%As shown in Fig.~\ref{fig:pointreg}, without introducing angle regression, PointReg is able to alleviate the boundary discontinuity problem. 
%Specifically, taking as input the feature maps extracted from backbone-FPN and the RRoI learned by the PointRPN, PointReg first generate RRoI features with a \textcolor{red}{RRoIAlign operation~\cite{han2021redet}. RRoIAlign~\cite{han2021redet} has shown effectiveness of performing RoI Align on rotated objects [TBD]}. Then, given the RRoI features, two consecutive fully-connected and ReLU layers are used to encode the RRoI features. Finally, two fully-connected layers are responsible for predicting the classification and location of the corresponding rotated object. 

\myparagraph{Optimization.}
The learning of PointReg is driven by classification loss and rotated object localization loss:
\begin{equation}
    \mathcal{L}_{PointReg} = \mu_1 \mathcal{L}_{cls} + \mu_2~ ^{+}\mathcal{L}_{loc},
\label{eq:pointreg_loss} 
\end{equation}
where $\mu_1$ and $\mu_2$ are the trade-off coefficients and are both set to 1.0 by default. $\mathcal{L}_{cls}$ indicates the classification loss, which is a cross-entropy loss:
\begin{equation}
    \mathcal{L}_{cls} = - \frac{1}{N} \sum_{i} \sum^{C}_{c=0} Y_{i \rightarrow c} \log (P_{i}),
\label{eq:pointreg_cls} 
\end{equation}
where $N$ denotes the number of training samples in PointReg, $C$ is the number of categories excluding background, $P_{i}$ is the predicted classification probability of the $i_{th}$ RRoI. $Y_{i \rightarrow c}=1$ if the ground-truth class of the $i_{th}$ RRoI is $c$; otherwise it is 0. 
$^{+}\mathcal{L}_{loc}$ represents the localization loss between the refined corners and the corners of ground-truth OBB. 
We use $L_1$ loss to optimize the corner refinement learning:
\begin{equation}
^{+}\mathcal{L}_{loc} = \frac{1}{N}\sum_{i}  \left | \mathcal{C}_i - \vartheta (b_i^*) \right |,
\end{equation}
where $\mathcal{C}_i (= \{(x_j, y_j)\}_{j=1}^{4} )$ denotes refined corners for the $i_{th}$ rotated proposal, $b_i^* (=\{ (x_j^*, y_j^*) \}_{j=1}^{4})$ denotes the corners of matched ground-truth OBB. $\vartheta (b_i^*)$ denotes the permutation of four corners of $b_i^*$ with the smallest $L_1$ loss $\left | \mathcal{C}_i - \vartheta (b_i^*) \right |$.
Note that $^{+}\mathcal{L}_{loc}$ is only calculated for positive training samples.

\subsection{Balanced Dataset}
\label{med:balancedataset}

The extremely nonuniform object densities of aerial images usually make the dataset long-tailed, which may cause the training process to be unstable and limit the detection performance.
For instance, DOTA-v1.0 contains 52, 516 ship instances but only 678 ground track field instances~\cite{ding2021object}.
%\textcolor{yellow}{To overcome this issue, in this section, we propose a balanced dataset strategy.}
To alleviate this issue, we resample the images of rare categories, which is inspired by~\cite{lvis2019}.
More concretely, first, for each category $c \in C$, compute the fraction of images $F_c$ that contains this category. Then, compute the category-level repeat factor for each category:
\begin{equation}
    r_c = \max (1.0, \sqrt{\beta_{thr} / F_c}),
\end{equation}
where $\beta_{thr}$ is a threshold which indicates that there will be not oversampling if ``$F_c > \beta_{thr}$''.
Finally, compute the image-level repeat factor $r_I$ for each image $I$:
\begin{equation}
    r_I = \underset{c \in C_I} \max (r_c),
\end{equation}
where $C_I$ denotes the categories contained in $I$.
In other words, those images that contain long-tailed categories will have a greater chance of being resampled during training.

\section{Experiment}

\subsection{Datasets}

To evaluate the effectiveness of our proposed Point RCNN framework, we perform experiments on two popular large-scale datasets: DOTA~\cite{xia2018dota} and HRSC2016~\cite{liu2017high}.
%and UCAS-AOD~\cite{zhu2015orientation}, 
%which are widely used for rotated object detection.

\textbf{DOTA}~\cite{xia2018dota} is the largest dataset for oriented object detection with three released versions: DOTA-v1.0, DOTA-v1.5 and DOTA-v2.0. To compare the performance with the state-of-the-art methods, we perform experiments on DOTA-v1.0 and DOTA-v1.5. DOTA-v1.0 contains 2806 images ranging in size from 800 $\times$ 800 to 4000 $\times$ 4000, and contains 188, 282 instances in 15 categories: Bridge (BR), Harbor (HA), Ship (SH), Plane (PL), Helicopter (HC), Small vehicle (SV), Large vehicle (LV), Baseball diamond (BD), Ground track field (GTF), Tennis court (TC), Basketball court (BC), Soccer-ball field (SBF), Roundabout (RA), Swimming pool (SP), and Storage tank (ST). DOTA-v1.5 has the same images as DOTA-v1.0 but contains 402, 089 instances. This is a more challenging dataset, which introduces a new category Container Crane (CC) and more small instances. 
%Same as previous methods~\cite{ding2019learning,han2021align,han2021redet}, we use training set and validation set for training. The performance are obtained by submitting prediction results to DOTA's evaluation server.

\textbf{HRSC2016}~\cite{liu2017high} %is another popular dataset for oriented object detection. It 
contains 1061 aerial images with size ranges from 300 $\times$ 300 to 1500 $\times$ 900. There are 436, 181, and 444 images in the training, validation and test set, respectively. 
%For a fair comparison, we use both the training and validation sets for training.

% UCAS-AOD~\cite{zhu2015orientation} %is another dataset for small oriented object detection, which 
% contains 1510 aerial images of about 659 $\times$ 1280 pixels, with 2 categories (car and plane) of 14,596 instances. For a fair comparison, same as UCAS-AOD-benchmark~\footnote{https://github.com/ming71/UCAS-AOD-benchmark}, we also divide the dataset into 755 images for training, 302 images for validation, and 453 images for testing with a ratio of 5:2:3.

\begin{table*}[t!]
\footnotesize 
\centering 
    \resizebox{1.0\textwidth}{!}{
    %\begin{tabular}{r|c|c|c|c|c|c|c|c|c|c|c|c|c|c|c|c|c}
    \begin{tabular}{r|c|ccccccccccccccc|c}
    \toprule
    Method                   & Backbone & PL & BD & BR & GTF & SV & LV & SH & TC & BC & ST & SBF & RA & HA & SP & HC & mAP (\%)  \\ \midrule
    RoI Trans.$^{*}$~\cite{ding2019learning} & R101-FPN & 88.64 & 78.52 & 43.44 & 75.92 & 68.81 & 73.68 & 83.59 & 90.74 & 77.27 & 81.46 & 58.39 & 53.54 & 62.83 & 58.93 & 47.67 & 69.56 \\ 
    O$^2$-DNet$^{*}$~\cite{wei2020oriented}  & H104 & 89.30 & 83.30 & 50.10 & 72.10 & 71.10 & 75.60 & 78.70 & 90.90 & 79.90 & 82.90 & 60.20 & 60.00 & 64.60 & 68.90 & 65.70 & 72.80 \\
    DRN$^{*}$~\cite{pan2020dynamic} & H104 & 89.71 & 82.34 & 47.22 & 64.10 & 76.22 & 74.43 & 85.84 & 90.57 & 86.18 & 84.89 & 57.65 & 61.93 & 69.30 & 69.63 & 58.48 & 73.23 \\
    Gliding Vertex$^{*}$~\cite{xu2020gliding} & R101-FPN & 89.64 & 85.00 & 52.26 & 77.34 & 73.01 & 73.14 & 86.82 & 90.74 & 79.02 & 86.81 & 59.55 & 70.91 & 72.94 & 70.86 & 57.32 & 75.02 \\ 
    BBAVectors$^{*}$~\cite{yi2021oriented}  & R101 & 88.63 & 84.06 & 52.13 & 69.56 & 78.26 & 80.40 & 88.06 & 90.87 & 87.23 & 86.39 & 56.11 & 65.62 & 67.10 & 72.08 & 63.96 & 75.36 \\
    CenterMap$^{*}$~\cite{wang2020learning} & R101-FPN & 89.83 & 84.41 & 54.60 & 70.25 & 77.66 & 78.32 & 87.19 & 90.66 & 84.89 & 85.27 & 56.46 & 69.23 & 74.13 & 71.56 & 66.06 & 76.03 \\
    CSL$^{*}$~\cite{yang2020arbitrary} & R152-FPN & 90.25 & 85.53 & 54.64 & 75.31 & 70.44 & 73.51 & 77.62 & 90.84 & 86.15 & 86.69 & 69.60 & 68.04 & 73.83 & 71.10 & 68.93 & 76.17 \\
    SCRDet++$^{*}$~\cite{yang2020scrdet++}  & R152-FPN & 88.68 & 85.22 & 54.70 & 73.71 & 71.92 & 84.14 & 79.39 & 90.82 & 87.04 & 86.02 & 67.90 & 60.86 & 74.52 & 70.76 & 72.66 & 76.56 \\
    CFA$^*$~\cite{BeyondBBox} & R-152 & 89.08 & 83.20 & 54.37 & 66.87 & 81.23 & 80.96 & 87.17 & 90.21 & 84.32 & 86.09 & 52.34 & 69.94 & 75.52 & 80.76 & 67.96 & 76.67 \\
    S$^{2}$A-Net$^{*}$~\cite{han2021align}  & R50-FPN & 88.89 & 83.60 & 57.74 & 81.95 & 79.94 & 83.19 & 89.11 & 90.78 & 84.87 & 87.81 & 70.30 & 68.25 & 78.30 & 77.01 & 69.58 & 79.42 \\
    ReDet$^{*}$~\cite{han2021redet} & ReR50-ReFPN & 88.81 & 82.48 & 60.83 & 80.82 & 78.34 & 86.06 & 88.31 & 90.87 & 88.77 & 87.03 & 68.65 & 66.90 & 79.26 & 79.71 & 74.67 & 80.10 \\
    Oriented R-CNN$^{*}$~\cite{xie2021oriented} & R101-FPN  & 90.26 & 84.74 & 62.01 & 80.42 & 79.04 & 85.07 & 88.52 & 90.85 & 87.24 & 87.96 & 72.26 & 70.03 & 82.93 & 78.46 & 68.05 & 80.52 \\
    Point RCNN$^{*}$ (Ours)         & ReR50-ReFPN & 82.99  & 85.73 & 61.16 & 79.98 & 77.82 & 85.90 & 88.94 & 90.89 & 88.89 & 88.16 & 71.84 & 68.21 & 79.03 & 80.32 & 75.71 & 80.37 \\
    
    Point RCNN$^{*\dag}$ (Ours)      & ReR50-ReFPN & 86.21 & 86.44 & 60.30 & 80.12 & 76.45 & 86.17 & 88.58 & 90.84 & 88.58 & 88.44 & 73.03 & 70.10 &  79.26 & 79.02 & 77.15  &  \textbf{80.71} \\
    
    Point RCNN$^{*\dag}$ (Ours)      & Swin-T-FPN  & 86.59 & 85.72 & 61.64 & 81.08 &  81.01 & 86.49 &  88.84 &  90.83 &  87.22 &  88.23 &  68.85 & 71.48 &  82.09 &  83.60 &  76.08  & \textbf{81.32} \\
    \bottomrule 
    \end{tabular}
}
\vspace{-.1in}
\caption{Performance comparisons on DOTA-v1.0 test set. $^*$ denotes multi-scale training and testing, $^{^*\dag}$ denotes the results of using balanced dataset strategy.}
\label{tbl:dota_1.0}
\end{table*}

\begin{table*}[t!]
\footnotesize 
\centering 
    \resizebox{1.0\textwidth}{!}{
    %\begin{tabular}{r|c|c|c|c|c|c|c|c|c|c|c|c|c|c|c|c|c|c}
    \begin{tabular}{r|c|cccccccccccccccc|c}
    \toprule
    Method   & Backbone & PL & BD & BR & GTF & SV & LV & SH & TC & BC & ST & SBF & RA & HA & SP & HC & CC & mAP (\%)  \\ \midrule
    RetinaNet-O~\cite{lin2017focal} & R50-FPN  & 71.43 & 77.64 & 42.12 & 64.65 & 44.53 & 56.79 & 73.31 & 90.84 & 76.02 & 59.96 & 46.95 & 69.24 & 59.65 & 64.52 & 48.06 & 0.83 & 59.16 \\
    FR-O~\cite{xia2018dota} & R50-FPN  & 71.89 & 74.47 & 44.45 & 59.87 & 51.28 & 68.98 & 79.37 & 90.78 & 77.38 & 67.50 & 47.75 & 69.72 & 61.22 & 65.28 & 60.47 & 1.54 & 62.00 \\
    Mask R-CNN~\cite{he2017mask} & R50-FPN  & 76.84 & 73.51 & 49.90 & 57.80 & 51.31 & 71.34 & 79.75 & 90.46 & 74.21 & 66.07 & 46.21 & 70.61 & 63.07 & 64.46 & 57.81 & 9.42 & 62.67 \\
    HTC~\cite{chen2019hybrid} & R50-FPN  & 77.80 & 73.67 & 51.40 & 63.99 & 51.54 & 73.31 & 80.31 & 90.48 & 75.12 & 67.34 & 48.51 & 70.63 & 64.84 & 64.48 & 55.87 & 5.15 & 63.40 \\
    OWSR$^{*}$~\cite{li2019learning} & R101-FPN  & - & - & - & - & - & - & - & - & - & - & - & - & - & - & - & - & 74.90 \\
    Oriented R-CNN$^{*}$~\cite{xie2021oriented} & R101-FPN  & 87.20 & 84.67 & 60.13 & 80.79 & 67.51 & 81.63 & 89.74 & 90.88 & 82.21 & 78.51 & 70.98 & 78.63 & 79.46 & 75.40 & 75.71 & 39.69 & 76.45 \\
    ReDet$^{*}$~\cite{han2021redet} & ReR50-ReFPN  & 88.51 & 86.45 & 61.23 & 81.20 & 67.60 & 83.65 & 90.00 & 90.86 & 84.30 & 75.33 & 71.49 & 72.06 & 78.32 & 74.73 & 76.10 & 46.98 & 76.80 \\
    %ReDet$^{*}$~\cite{han2021redet} & Swin-T-FPN  & 80.90 & 85.13 & 60.61 & 80.83 & 67.07 & 83.32 & 89.80 & 90.79 & 86.04 & 78.69 & 75.35 & 77.38 & 78.48 & 75.41 & 79.51 & 61.95 & 78.20 \\
    Point RCNN$^{*}$ (Ours) & ReR50-ReFPN  & 83.40 & 86.59 & 60.76 & 80.25 & 79.92 & 83.37 & 90.04 & 90.86 & 87.45 & 84.50 & 72.79 & 77.32 & 78.29 & 77.48 & 78.92 & 47.97 & 78.74 \\
    Point RCNN$^{*\dag}$ (Ours) & ReR50-ReFPN  & 83.12 & 86.55 & 60.84 & 82.43 & 80.60 & 83.39 & 90.01 & 90.88 & 87.25 & 84.60 & 73.49 & 78.51 & 78.75 & 78.41 & 76.12 & 54.12 & \textbf{79.31} \\
    %Point RCNN$^{*}$ (Ours) & Swin-T-FPN  & 83.88 & 85.22 & 60.76 & 79.40 & 81.64 & 83.48 & 89.98 & 90.75 & 87.00 & 84.65 & 70.70 & 77.87 & 78.32 & 79.50 & 74.35 & 63.80 & \textbf{79.46} \\
    Point RCNN$^{*\dag}$ (Ours) & Swin-T-FPN  & 86.93  & 85.79 & 59.52 & 80.42 & 81.91 & 81.92 & 89.95 & 90.35 & 85.72 & 85.84 & 68.57 & 76.35 & 78.79 & 81.24 & 78.64 & 69.23  & \textbf{80.14} \\
    \bottomrule
    \end{tabular}  
}
\vspace{-.1in}
\caption{Performance comparisons on DOTA-v1.5 test set. $^*$ denotes multi-scale training and testing, $^{^*\dag}$ denotes the results of using balanced dataset strategy. 
%Note that the results of Faster R-CNN OBB (FR-O)~\cite{xia2018dota}, RetinaNet OBB (RetinaNet-O)~\cite{lin2017focal}, Mask R-CNN~\cite{he2017mask} and Hybrid Task Cascade (HTC)~\cite{chen2019hybrid} are excerpted from ReDet~\cite{han2021redet}. The results of Oriented R-CNN$^{*}$ and ReDet$^{*}$ with Swin-T-FPN backbone are our re-implementations based on their official code.
}
\label{tbl:dota_1.5}
\vspace{-.1in}
\end{table*}

\subsection{Implementation Details}

We implement Point RCNN using the MMDetection tool-box~\cite{mmdetection}. 
%For a fair comparison with ReDet~\cite{han2021redet}, 
We follow ReDet~\cite{han2021redet} to use ReResNet with ReFPN as our backbone (ReR50-ReFPN), which has shown the ability to extract rotation-equivariant features. 
We also verify with the more generalized transformer backbone (Swin-Tiny) to show the generalization and scalability of our Point RCNN.
%and the model weights are initialized by the pretrained model of ReDet
%, and use Rotation-invariant RoI Align (RiRoI Align) as the RoI operation.
%Rotation-equivariant networks have shown the ability to extract rotation-equivariant features.
%and RiRoI Align can extract rotation-invariant features from the rotated RoI.

On the DOTA dataset, following previous methods~\cite{ding2019learning,han2021align,han2021redet}, we crop the image to 1024 $\times$ 1024 with 824 pixels as a stride and we also resize the image to three scales $\{0,5, 1.0, 1.5\}$ for multi-scale data. Random horizontal flipping and random rotation ($[-45{^\circ}, 45{^\circ}]$) are adopted for multi-scale training. 
On the HRSC2016 dataset, like previous method~\cite{han2021redet}, we resize all the images to (800, 512), random horizontal flipping is applied during training.
%
%On the UCAS-AOD dataset, following the UCAS-AOD-benchmark, we resize all the images to (800, 800) and only use the training set for training. %Random horizontal flipping, HSV augment and random rotation are applied during training.
%
Unless otherwise specified, we train all the models with 19 epochs for DOTA and 36 epochs for HRSC2016.
%and UCAS-AOD. 
To be specific, we train the models using AdamW~\cite{adam} on 8 Tesla-V100 GPUs with $\beta_1$=0.9 and $\beta_2$=0.999, with an initial learning rate of 0.0002, a weight decay of 0.05, and a mini-batch size of 16 (2 images per GPU). The learning rate decays by a factor of 10 at each decay step.

\begin{table}[t!]
\footnotesize 
\centering
\resizebox{0.8\linewidth}{!}{
    \begin{tabular}{@{}r|c|c|c@{}}
    \toprule
    Method & Backbone & mAP$_{\text{07}}$ (\%) & mAP$_{\text{12}}$ (\%)  \\ 
    \midrule
    RC2~\cite{liu2017learning} & VGG16 & 75.70 & - \\
    RRPN~\cite{ma2018arbitrary} & R101  & 79.08 & 85.64 \\
    R$^{2}$PN~\cite{zhang2018toward} & VGG16 & 79.60 & - \\
    RRD~\cite{liao2018rotation} & VGG16 & 84.30  & - \\
    RoI-Trans.~\cite{ding2019learning} & R101-FPN & 86.20  & -  \\
    Gliding Vertex~\cite{xu2020gliding} & R101-FPN & 88.20 & -  \\
    R$^{3}$Det~\cite{yang2019r3det} & R101-FPN & 89.26  & -  \\
    DRN~\cite{pan2020dynamic} & H34  & -  &  92.7 \\
    CenterMap~\cite{wang2020learning} & R50-FPN & -  & 92.8 \\
    CSL~\cite{yang2020arbitrary}  & R152-FPN  & 89.62  & - \\
    S$^{2}$A-Net~\cite{han2021align} & R101-FPN & 90.17  &  95.01 \\
    ReDet~\cite{han2021redet} & ReR50-ReFPN & 90.46  &  97.63  \\
    Orient R-CNN~\cite{xie2021oriented} & R101-FPN & 90.50 & 97.60 \\
    Point RCNN (Ours) & ReR50-ReFPN & \textbf{90.53} & \textbf{98.53} \\
    \bottomrule
    \end{tabular}
}
\vspace{-.05in}
\caption{Performance comparisons on HRSC2016 test set. mAP$_{\text{07}}$ and mAP$_{\text{12}}$ indicate that the results are evaluated under VOC2007 and VOC2012 metrics(\%) respectively. We report both results for fair comparisons.}
\label{tbl:hrsc2016}
\vspace{-.1in}
\end{table}

\subsection{Main Results}

We compare our Point RCNN framework with other state-of-the-art methods on three datasets: DOTA-v1.0, DOTA-v1.5, and HRSC2016.
%and UCAS-AOD. 
As shown in~\cref{tbl:dota_1.0},~\cref{tbl:dota_1.5}, and~\cref{tbl:hrsc2016}, %and~\cref{tbl:ucas_aod}, 
without bells and whistles, our Point RCNN demonstrates superior performance against state-of-the-art methods. 

\textbf{On DOTA-v1.0}, as reported in~\cref{tbl:dota_1.0}, Point RCNN achieves state-of-the-art \textbf{80.71} mAP. With the more generalized transformer backbone Swin-Tiny~\cite{liu2021swin} (Swin-T), Point RCNN can further improve the performance by \textbf{0.61\%} (from 80.71 to 81.32).

\textbf{On DOTA-v1.5}, which is more challenging compared to DOTA-v1.0, Point RCNN achieves \textbf{79.31} mAP, which significantly improve the performance by \textbf{2.51\%}.
With the more generalized transformer backbone Swin-T, Point RCNN further improves the performance by \textbf{0.83\%} (from 79.31 to 80.14).
The results are reported in~\cref{tbl:dota_1.5}

\textbf{On HRSC2016}, as reported in~\cref{tbl:hrsc2016}, Point RCNN attains the new state-of-the-art performance under both the VOC2007 and VOC2012 metrics, respectively.

\subsection{Ablation Study}

In this section, if not specified, all the models are trained only on the training and validation set with scale 1.0 for simplicity, and are tested using multi-scale testing. The metric mAP is evaluated on the DOTA-v1.5 test set and obtained by submitting prediction results to DOTA's evaluation server.

%\subsubsection{Ablation on PointRPN Design}
%
% \begin{table}[!h]
% \small
% \centering
% \begin{tabular}{@{}c|c|c@{}}
% \toprule
% CIoU Loss & Vertex L1 Loss & mAP (\%) \\ \midrule
% $\checkmark$     &                 & 74.17                 \\ %\midrule
% $\checkmark$     & $\checkmark$   &  74.62     \\ \bottomrule  
% \end{tabular}
% \caption{\textcolor{red}{Change results to quad head.} Comparison of OBB detection performance on the test-dev set of DOTA V1.5 dataset. \textit{CIoU} denotes Convex-hull loss used in \cite{BeyondBBox}. \textit{Vertex L1} represents the L1 loss applied on the first four points of the prediction. The first four points predicted should correspond to the four vertices of the corresponding ground-truth rotated instance.}
% \end{table}

% \begin{table}[!h]
% \small
% \centering
% \begin{tabular}{@{}c|c|c@{}}
% \toprule
% \multicolumn{2}{l|}{Label Assignment} & \multirow{2}{*}{mAP (\%)} \\ \cmidrule(r){1-2}
% MaxIoU             & Topk               &              \\ \midrule
%     &   &         \\ \bottomrule  
% \end{tabular}
% \caption{Comparison of OBB detection performance on the test-dev set of DOTA V1.5 dataset.}
% \end{table}

% \subsubsection{Ablation on Corner Heatmap}
%
% \begin{table}[!h]
% \small
% \centering
% \begin{tabular}{@{}c|c@{}}
% \toprule
% Vertex Feature Fusion & mAP (\%) \\ \midrule
%                       &  73.52  \\ 
%   $\checkmark$        &  74.02   \\ \bottomrule
% \end{tabular}
% \caption{Comparison of OBB detection performance on the test-dev set of DOTA V1.5 dataset.}
% \end{table}  

\subsubsection{Effect of PointRPN}

To analysis the efficiency of PointRPN, we evaluate the detection recall of PointRPN on the validation set of DOTA-v1.5.
For simplicity, we train the models on the training set with scale 1.0 and evaluate the metric recall on the validation set with scale 1.0 as well. The positive IoU threshold is set to 0.5. 
We select the top-300, top-1000, and top-2000 proposals to calculate their recall values and report the results in~\cref{tbl:pointrpn_recall}.
We can find that when the proposals changes from top-2000 to top-1000, the recall value only drops 0.17\%. 
Even if there are only top-300 proposals, the recall still achieves 85.93\%.

\begin{table}[!h]
\small
\centering
\resizebox{0.8\linewidth}{!}{
\begin{tabular}{@{}r|c|c|c@{}}
\toprule
Method      & Recall$_{300}$ (\%)  & Recall$_{1000}$ (\%) & Recall$_{2000}$ (\%)  \\ \midrule
PointRPN    & 85.93 & 89.83 & 90.00 \\ 
\bottomrule
\end{tabular}
}
\vspace{-.05in}
\caption{Comparison of the recall results by varying the number of proposals of each image patch. The metric recall is evaluated on the DOTA-v1.5 validation set.}
\label{tbl:pointrpn_recall}
\vspace{-.1in}
\end{table}

\begin{figure}[t!]
\centering 
\includegraphics[width=0.95\linewidth]{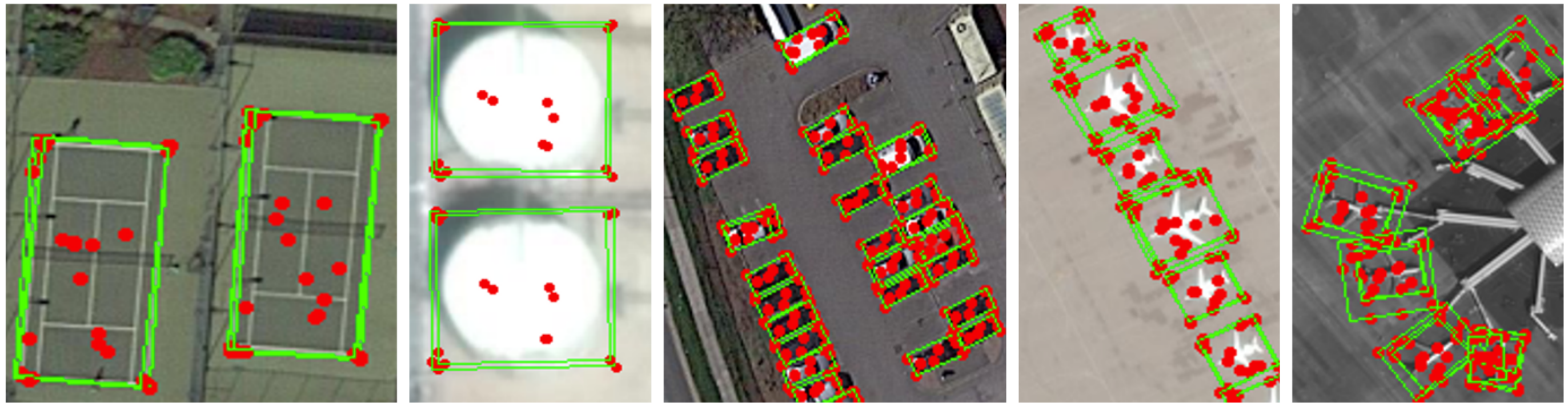}
\vspace{-.1in}
\caption{Examples of the learned points (red points) of PointRPN on DOTA-v1.0 test set. Green OBBs are the converted pseudo OBBs. The score threshold is set to 0.001 without NMS.}
\label{fig:vis_rpn}
\vspace{-.15in}
\end{figure}
 
\begin{figure*}[!t]
\centering 
\includegraphics[width=0.9\linewidth]{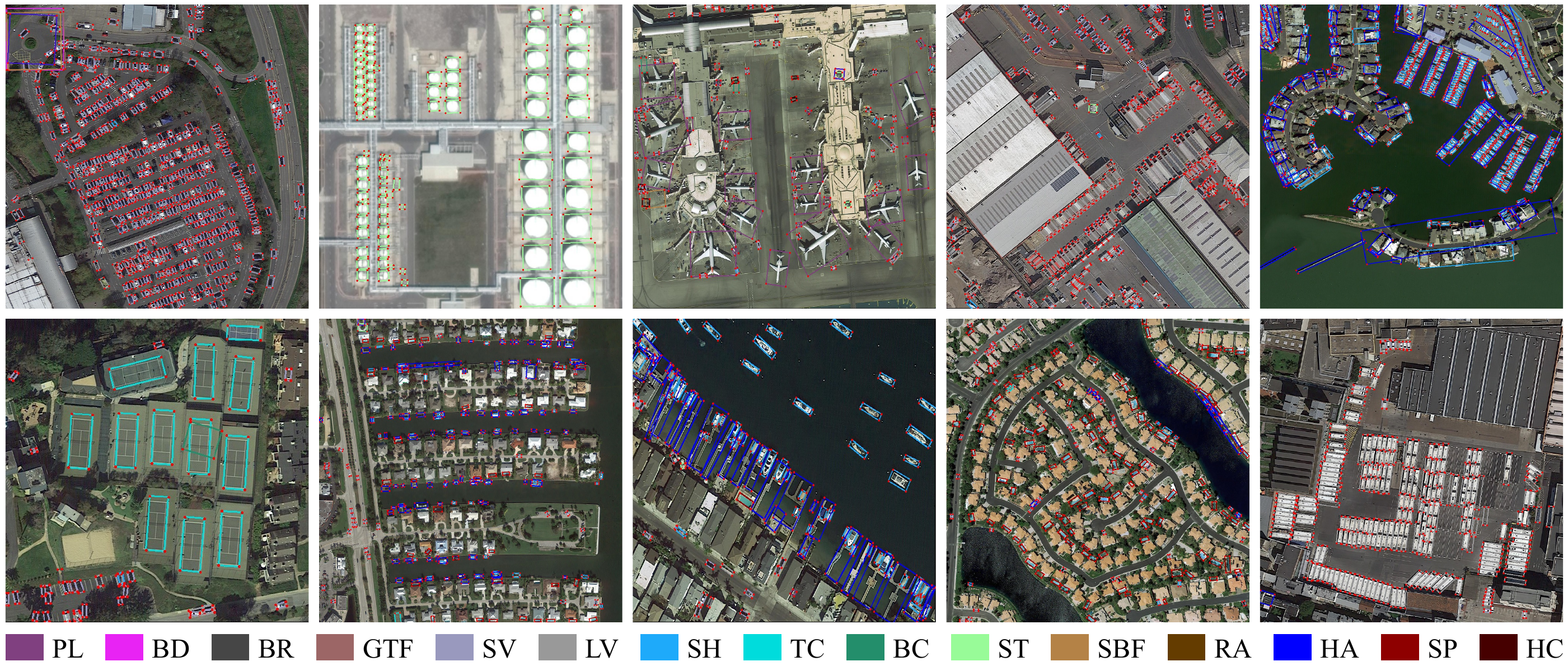}
\vspace{-.1in}
\caption{Visualization of the detection results of Point RCNN on DOTA-v1.0 test set. The score threshold is set to 0.01. Each color represents a category. The red points and colored OBBs are the predicted corner points and the converted OBBs of PointReg.}
\label{fig:vis_rcnn}
\vspace{-.1in}
\end{figure*}

\subsubsection{Effect of Regression Type of PointReg}

In this section, we analysis the effectiveness of the OBB regression type of PointReg. The results are shown in~\cref{tbl:regress_type}, compared to the five-parameter representation, the eight-parameter regression type brings higher performance.

\begin{table}[!h]
\small
\centering
\resizebox{0.6\linewidth}{!}{
\begin{tabular}{@{}c|c@{}}
\toprule
Regression type & mAP (\%)   \\ \midrule
($x, y, w, h, \theta$)    & 77.25 \\ 
($x_1, y_1, x_2, y_2, x_3, y_3, x_4, y_4$)    & \textbf{77.60}    \\ 
\bottomrule
\end{tabular}
}
\vspace{-.05in}
\caption{Analysis of the effectiveness of OBB regression type of PointReg. The metric mAP is evaluated on the DOTA-v1.5 test set.}
\label{tbl:regress_type}
\vspace{-.1in}
\end{table}

%%%%%% UCAS-AOD
% \begin{table}[t!]
% \footnotesize 
% \centering
% \resizebox{0.9\linewidth}{!}{
%     \begin{tabular}{r|c|c|c|c}
%     \toprule
%     Method & Backbone & car (\%) & airplane (\%) & mAP (\%)  \\ 
%     \midrule
%     R-Yolov3~\cite{redmon2018yolov3}  & Darknet53 & 74.63 & 89.52 & 82.08 \\
%     R-RetinaNet~\cite{lin2017focal}	& ResNet50 & 84.64 & 90.51 & 87.57 \\
%     RoI-Trans.~\cite{ding2019learning} & ResNet50 & 88.02 &	90.02 & 89.02 \\
%     DAL~\cite{ming2020dynamic} & ResNet50 & 89.25 & 90.49 & 89.87 \\
%     S$^{2}$A-Net~\cite{han2021align} & ResNet50 & 89.56 & 90.42 & 89.99 \\
%     Point RCNN (Ours) & ReR50-ReFPN & 89.60 & 90.48 & \textbf{90.04} \\
%     \bottomrule
%     \end{tabular}
% }
% \caption{Performance comparisons on UCAS-AOD test set. All models are evaluated via VOC2007 metric (\%).}
% \label{tbl:ucas_aod}
% \end{table}

\subsubsection{Effect of Balanced Dataset}
\begin{table}[!h]
\small
\centering
\resizebox{0.65\linewidth}{!}{
\begin{tabular}{@{}c|c@{}}
\toprule
Oversampling threshold ($\beta_{thr}$) & mAP (\%)   \\ \midrule
0      & 73.52 \\ 
0.1    & 76.49 \\
0.2    & 77.44 \\ 
0.3    & \textbf{77.60} \\
0.4    & 77.48 \\ \bottomrule
\end{tabular}
}
\vspace{-.1in}
\caption{Comparison of detection accuracy by varying the oversampling threshold $\beta_{thr}$. The metric mAP is evaluated on the DOTA-v1.5 test set.}
\label{tab:balanced}
\vspace{-.1in}
\end{table}

In this section, we analysis the impact of the oversampling threshold $\beta_{thr}$ of the balanced dataset strategy. As shown in~\cref{tab:balanced}, we achieve the best detection accuracy of 77.60\% at $\beta_{thr} = 0.3$. Therefore, we set $\beta_{thr} = 0.3$ in all other experiments on DOTA.

\subsubsection{Factor-by-factor Experiment}

\begin{table}[!h]
\small
\centering
\resizebox{0.9\linewidth}{!}{
    \begin{tabular}{@{}r|c|c|c|c@{}}
    \toprule
    Method   & PointRPN & Balanced Dataset & PointReg & mAP (\%) \\ \midrule
    Baseline &          &                &          &  71.36   \\
    \midrule
    \multirow{4}{*}{Point RCNN}         &  $\checkmark$ &                &          &  74.17   \\
                                        &          &     $\checkmark$    &          &  74.22   \\
                                        
                                        &  $\checkmark$ &  $\checkmark$ &          &  77.25   \\
                                        &  $\checkmark$ &  $\checkmark$ &  $\checkmark$  &  \textbf{77.60} \\
    \bottomrule
    \end{tabular}
}
\vspace{-.1in}
\caption{Factor-by-factor ablation experiments. The detection performance is evaluated on the test set of DOTA-v1.5 dataset.}
\label{tbl:factorbyfactor}
\end{table}

To explore the effectiveness of each module of our proposed Point RCNN framework, we conduct a factor-by-factor experiment on the proposed PointRPN, PointReg and balanced dataset strategy. The results are depicted in~\cref{tbl:factorbyfactor}, each component has a positive effect, and all components are combined to obtain the best performance.

\subsubsection{Visualization Analysis}

We visualize some detection results on DOTA-v1.0 test set. Fig.~\ref{fig:vis_rpn} shows some examples of the learned points of PointRPN, which indicates that PointRPN is capable of learning representative points of rotated object.
Fig.~\ref{fig:vis_rcnn} shows some final detection results of Point RCNN, the red points denote the corner points learned by PointReg and the colored OBBs converted by the $\tt{MinAreaRect}$ function of OpenCV are the final results.

\subsection{Limitations}

Although experiments substantiate the superiority of Point RCNN over state-of-the-art methods, our method does not perform well enough in some categories, \eg, PL (Plane), which needs to be further explored.
Point RCNN also needs to use rotate NMS to remove duplicate results, which may mistakenly delete the TP.
Transformer-based methods~\cite{DETR} may be the potential solutions, which will be the future work.

%%%%%%%%% Conclusion
\section{Conclusion}

In this work, we revisit rotated object detection and propose a purely angle-free framework for rotated object detection, named Point RCNN, which mainly consists of a PointRPN for generating accurate RRoIs, and a PointReg for refining corner points based on the generated RRoIs. 
In addition, we propose a balanced dataset strategy to overcome the long-tailed distribution of different object classes in aerial images. 
Extensive experiments on several large-scale benchmarks demonstrate the significant superiority of our proposed framework against the state-of-the-arts.

%%%%%%%%% REFERENCES
{\small
\bibliographystyle{ieee_fullname}
\bibliography{egbib}
}

\end{document}